\documentclass[conference]{IEEEtran}
\IEEEoverridecommandlockouts
\usepackage[utf8]{inputenc} 
\usepackage[T1]{fontenc}    
\usepackage{url}            
\usepackage{booktabs}       
\usepackage{amsfonts}       
\usepackage{nicefrac}       
\usepackage{microtype}      
\usepackage{xcolor}         

\usepackage{times}
\usepackage{soul}
\usepackage{graphicx}
\usepackage{amsthm,mathrsfs}
\usepackage{verbatim} 
\usepackage{listings} 
\usepackage{algorithm}
\usepackage{algorithmic}
\usepackage{siunitx}
\usepackage{bm}
\usepackage{overpic}
\usepackage{fontawesome}

\usepackage{amsmath,amssymb}
\usepackage{textcomp}
\usepackage{multirow}
\usepackage{pgfplots}
\pgfplotsset{compat=newest}
\usepackage{subcaption}
\usepackage{tikz,pgf,tikzscale}
\usepackage{tikz}
\usetikzlibrary{shapes,patterns,shapes.geometric,arrows.meta,bending,positioning,graphs,calc}
\usepackage{pgf-pie}
\usepackage{transparent}
\usepackage{standalone}

\newcommand{\iou}{\mathit{IoU}}

\newcommand{\map}{\mathit{mAP}}

\usepackage[pagebackref,breaklinks,colorlinks]{hyperref}
\def\BibTeX{{\rm B\kern-.05em{\sc i\kern-.025em b}\kern-.08em
    T\kern-.1667em\lower.7ex\hbox{E}\kern-.125emX}}

\makeatletter
\newcommand{\linebreakand}{%
  \end{@IEEEauthorhalign}
  \hfill\mbox{}\par
  \mbox{}\hfill\begin{@IEEEauthorhalign}
}
\makeatother

\usepackage[capitalize]{cleveref}

\begin{document}

\title{Deep Active Learning with Noisy Oracle in Object Detection}

\author{
    Marius Schubert${}^{\ast 1}$ \hspace{2em}
    Tobias Riedlinger${}^{\ast 2}$ \hspace{2em}
    Karsten Kahl${}^1$ \hspace{2em}
    Matthias Rottmann${}^1$
    \\[0.5em]
    ${}^1$University of Wuppertal, School of Mathematics and Natural Sciences, IZMD \\
    \{schubert, kkahl, rottmann\}@math.uni-wuppertal.de \\[0.5em]
    ${}^2$Technical University of Berlin, Institute of Mathematics  \\
    riedlinger@tu-berlin.de
}

\maketitle

\def\thefootnote{$\ast$}\footnotetext{Equal contribution.}

\begin{abstract}
Obtaining annotations for complex computer vision tasks such as object detection is an expensive and time-intense endeavor involving a large number of human workers or expert opinions.
Reducing the amount of annotations required while maintaining algorithm performance is, therefore, desirable for machine learning practitioners and has been successfully achieved by active learning algorithms.
However, it is not merely the amount of annotations which influences model performance but also the annotation quality.
In practice, the oracles that are queried for new annotations frequently contain significant amounts of noise.
Therefore, cleansing procedures are oftentimes necessary to review and correct given labels.
This process is subject to the same budget as the initial annotation itself since it requires human workers or even domain experts.
Here, we propose a composite active learning framework including a label review module for deep object detection.
We show that utilizing part of the annotation budget to correct the noisy annotations partially in the active dataset leads to early improvements in model performance, especially when coupled with uncertainty-based query strategies.
The precision of the label error proposals has a significant influence on the measured effect of the label review.
In our experiments we achieve improvements of up to $4.5$ $\map$ points of object detection performance by incorporating label reviews at equal annotation budget.
\end{abstract}


\section{Introduction}
\begin{figure}
    \centering
    \resizebox{.99\linewidth}{!}{
    \input{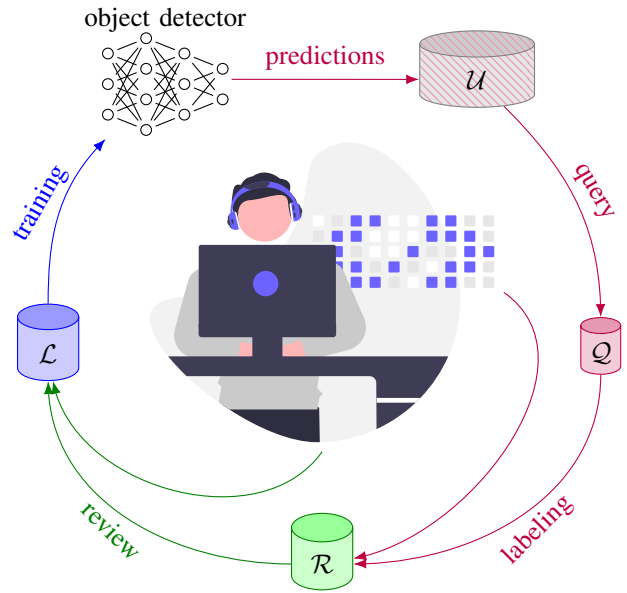}
    }
    \caption{Our active learning cycle consists of {\color{blue}training} on labeled data $\mathcal{L}$, {\color{purple}querying and labeling} informative data points $\mathcal{Q}$ out of a pool of unlabeled data $\mathcal{U}$ by an oracle and a {\color{green!50!black}review} of acquired data $\mathcal{R}=\mathcal{L}\cup\mathcal{Q}$.
    }
    \label{fig: al cycle}
\end{figure}
\noindent
In the previous decade, deep learning has revolutionized computer vision models across many different tasks like supervised object detection~\cite{ren2015faster,redmon2018yolov3,carion2020end}.
Object detection has various potential real-world applications, many of which have not yet been developed in a sense that public datasets are rarely or just not available.
When such a new field is to be developed, there are many practical challenges during dataset curation and creation. 
Oftentimes, data can be recorded with, e.g., cameras in large amount at acceptable cost, while acquiring corresponding labels might be comparatively costly and might require expert knowledge. 
Active learning aims at maintaining model performance while reducing the amount of training data by leveraging data informativeness for the label acquisition.
The model is utilized in turn to find the data, in our case from a large pool of unlabeled data, for which new labels would improve the model performance most efficiently~\cite{settles2009active,brust2018active,riedlinger2022activelearning}. 
The goal is to request as few annotations with human labor as possible and to obtain a well-performing model that makes accurate predictions. 
When developing and simulating active learning models in a laboratory setup, one typically assumes an oracle that provides correct labels for queried data points~\cite{brust2018active,riedlinger2022activelearning}. 
However, in practice, such an oracle does not exist and any person that labels data produces errors with some frequency~\cite{yan_learning_2014}. 
Especially in complex domains such as medical applications where expert opinions are required for the annotation process, there exists variability between different oracles~\cite{schilling_automated_2022}.
Some works have considered active learning with noisy oracles in image classification~\cite{younesian_active_2020,younesian_qactor_2021,yan_active_2016,gupta_noisy_2019}. 
In the present work, we consider active learning with a noisy oracle, to the best of our knowledge for the first time, in object detection. 
More precisely, we utilize recent findings on label errors in state-of-the-art object detection benchmarks~\cite{schubert2023identifying} to simulate two types of predominantly occurring label errors in object detection oracles. 
On the one hand we treat missed bounding box labels which do not appear in the ground truth at all.
On the other hand, we consider bounding box labels with incorrect class assignment which are likely to induce undesired training feedback. 
We do so for the EMNIST-Det dataset~\cite{riedlinger2022activelearning} which is an extension of EMNIST~\cite{cohen2017emnist} to the object detection and instance segmentation setting.
We complement this with the BDD100k dataset~\cite{yu2020bdd100k} which has mostly clean bounding box annotations of variable size. 
Both datasets have high quality bounding box labels such that we can simulate label errors without greater influence of naturally occurring label noise. 
We introduce independent and identically distributed errors into the labels which have been queried at some point in the data-acquisition process. 
We simulate a label reviewer as a human in the loop who has access to a label error detection module~\cite{schubert2023identifying} which is integrated into the active learning cycle, see \cref{fig: al cycle}. 
We compare different sources for label error proposals which are to be considered after data acquisition.
We use different methods to generate label error proposals for the reviewer.
The efficiency of the proposal method controls the frequency of justified review cases, i.e., the efficiency of the budget utilization for label reviewing.
The review oracle is assumed to contain smaller amounts of noise since labels do not have to be generated from scratch. 
Instead, only individual proposals have to be reviewed.


In our experiments, we observe that a label error detection method applied to active learning with a noisy oracle clearly outperforms active learning with random label review and active learning without any label review.
We compare different query strategies with and without review in terms of performance as a function of annotation budget (split into labeling and reviewing cost). 
Improvement of performance is observed consistently for random queries as well as for an uncertainty query based on the entropy of the object detector's softmax output. 
Furthermore, our findings are consistent over two datasets, i.e., an artificial one and a real world one, as well as across two different object detectors. 
The success of our method seems to be due to a strong performance of the label error detection method.

Our contribution can be summarized as follows:
\begin{itemize}
    \item We contribute the first method that performs partially automated label review and active label selection for object detection.
    \item We provide an environment for performing rapid prototyping of methods for active learning with noisy oracles.
    \item Our method outperforms manual and review-free active learning for different queries, datasets and underlying object detectors.
\end{itemize}
Our method can be used with humans in the loop for labeling and label review to maximize model performance at minimal annotation budget, thus aiding data acquisition pipelines with partial automation.

\section{Related Work}
\noindent
Our contribution is located at the intersection of two disciplines which both aim at reducing the tiresome workload of repetitive image annotation by human workers.
Active learning aims at reducing the overall amount of annotations given while the goal of label reviewing is to control or improve the quality of present annotations.

\paragraph{Label Error Detection in Object Detection}
Previous work on the detection of label errors for object detectors tend to make use of a model which was trained on given, potentially error-prone data.
Hu et al.~\cite{hu_probability_2022} compare a softmax probability-based measure per prediction with the given annotations to obtain proposals for label errors.
Schubert et al.~\cite{schubert2023identifying} used an instance-wise loss computation to identify different types of label errors.
Koksal et al.~\cite{koksal_effect_2020}, in contrast, use a template matching scheme to find annotation errors in frame sequences for UAV detection.

\paragraph{Active Learning in Object Detection}
Training in the context of more refined computer vision tasks such as object detection requires significant computational resources and the learning task itself comes with an elevated degree of complexity.
All the more important is efficient handling when it comes to expensive data annotations which can be approached by active learning.
The following accounts for the pioneering successes that were accomplished in fully supervised active learning for deep object detection.
Brust et al.~\cite{brust2018active} used margin sampling by aggregating probability margin scores over predicted bounding boxes in different ways where a class-weighting scheme addresses class imbalances.
Roy et al.~\cite{roy_deep_2018} similarly utilize softmax entropy and committee-based scoring making use of the different detection scales of object detection architectures.
Schmidt et al.~\cite{schmidt_advanced_2020} compare different deep ensemble consensus-based selection strategies leveraging model uncertainty.
Choi et al.~\cite{choi_active_2021} utilize Monte-Carlo dropout in conjunction with a Gaussian mixture model to estimate epistemic and aleatoric uncertainty, respectively.
The utilized scoring function for querying takes both kinds of uncertainty into account.
Haussmann et al.~\cite{haussmann_scalable_2020} compare methodically diverse scoring functions in the objectness entropy, mutual information, estimated learning gradient and confidence coupled with different diversity selection mechanisms.

\paragraph{Active Learning with Noisy Oracle in Classification}
The intersection between active learning and training under label noise has been addressed in the context of classification tasks before.
While Kim~\cite{kim_active_2022} used an active query mechanism for cleaning up labels, the proposed training algorithm itself is not active.
Younesian et al.\ \cite{younesian_active_2020} consider noisy binary and categorical oracles by assigning different label costs to both in an online, stream-based active learning setting.
Yan et al.\ \cite{yan_active_2016} treat the query complexity of noisy oracles with a reject option in a theoretical manner.
Gupta et al.\ \cite{gupta_noisy_2019} consider batch-based active learning with noisy oracles under the introduction of the QActor framework by Younesian et al.\ \cite{younesian_qactor_2021} which has a label cleaning module in its active learning cycle is most related to our approach.
One of the proposed quality models chooses examples to clean via the cross-entropy loss which are then re-labeled by the oracle.


\section{Active Learning with Noisy Oracle}
\noindent
In this section, we describe the task of active learning in object detection as well as the addition of a review module to the generic active learning cycle.
While in the active learning setting, new labels are queried on the basis of an informativeness measure, the presence of erroneous or misleading oracle responses can counteract the benefit of the informed data selection.
In order to account for new data containing incorrect labels, we introduce a review module that generates proposals for label errors to review and to potentially correct.
\subsection{Active Learning with Review in Object Detection}
\noindent
Most of the commonly used datasets in object detection, e.g., MS-COCO~\cite{lin2014microsoft} and Pascal VOC~\cite{everingham2010pascal}, are also the most commonly used datasets in active learning and contain label errors~\cite{schubert2023identifying,rottmann2023automated}. 
This means that active learning methods developed on these datasets are also evaluated based on noisy labels.
To consider label errors during active learning experiments, we introduce a review module.
Active learning can be viewed as an alternating process of training a model and labeling data based on informativeness according to the model.
Starting with an initially labeled set of images $\mathcal{L}$, once a model is trained based on $\mathcal{L}$, the test performance is measured.
Object detectors are usually evaluated in terms of mean average precision ($\map$, see~\cite{everingham2010pascal}).
New images selected to be labeled ($\mathcal{Q}$) are queried from a pool $\mathcal{U}$ of previously unlabeled images.
After obtaining labels for the queried images $\mathcal{Q}$ by an oracle, we introduce a review step where an oracle reviews $\mathcal{R}=\mathcal{L}\cup\mathcal{Q}$.
The model is then trained on the reviewed data and the cycle is repeated $T$ times.
The active learning cycle is visualized in \cref{fig: al cycle} where the annotation budget is divided into the query and the review with parameter $\lambda\in[0,1]$.
Note, that acquisition and review of data are two independent modules.

\paragraph{Queries}
Active learning research usually revolves around the development of model architectures, loss functions or selection strategies used in the query step.
Different query approaches are then compared for different annotation budgets in terms of the achieved test performance which is often measured in terms of $\map$ in object detection.
In the following, we investigate two different query strategies: random selection and selection based on the entropy of the softmax output of the object detector. 
For the former, images are randomly chosen from $\mathcal{U}$.
For the latter, images are selected based on the predictive classification uncertainty according to the current model.
For a given image $x$, a neural network predicts a fixed number $N_0$ of bounding box predictions
\begin{equation}
    \hat{b}^i=\{(\hat{x}^i,\hat{y}^i,\hat{w}^i,\hat{h}^i,\hat{s}^i,\hat{p}_1^i,\ldots,\hat{p}_C^i)\},
\end{equation}
where $i=1,\ldots,N_0$.
Here, $\hat{x}^i,\hat{y}^i,\hat{w}^i,\hat{h}^i$ represent the localization, $\hat{s}^i$ the objectness score and $\hat{p}_1,\ldots,\hat{p}_C$ the class probabilities for all classes $\{1,\ldots,C\}$.
Only the set of boxes that remain after score-thresholding (with threshold $s_\epsilon$) and non-maximum suppression (NMS) are used to compute prediction-wise entropies.  
The entropy $H(\hat{b}^i)$ of a prediction $\hat{b}^i$ is given by $H(\hat{b}^i)=-\sum_{c=1}^C \hat{p}_c^i\cdot\text{log}(\hat{p}_c^i)$.  
Moreover, we incorporate a class-weighting~\cite{brust2018active} for computing instance-wise uncertainty scores.
Finally, the instance-wise entropies of a single image are summed up to obtain an image-wise query score and the images are sorted in descending order by this query score.
Note, that both queries are independent of the label errors, since the random selection is independent of predictions and labels as well as the image-wise query score for the entropy method is determined based on the predictions only. 
Note, that query algorithms are based on unlabeled data and oracle noise does not directly impact the selection of images.
However, oracle noise does influence model training.


\paragraph{Review Module}
In order to account for noisy oracles, we allow for incorrect annotations given in response to an active learning query. 
To counter-act the negative influence of noisy annotations, we introduce a review module into the active learning cycle in which proposals for label reviews are given and part of the annotation budget is used to clean up some of the annotation errors.
In the following, we introduce the detection of two different kinds of label errors: missing labels (\emph{misses}) and labels with incorrect class assignments (\emph{flips}).

For one active learning cycle, we allow for the consumption of a fixed annotation budget $\mathcal{C}$.
This budget $\mathcal{C}$ is split up into a fraction $\mathcal{C}_\mathcal{Q} = (1 - \lambda) \mathcal{C}$ used for querying new annotations and $\mathcal{C}_\mathcal{R} = \lambda \mathcal{C}$ used for reviewing data.

After the query, $\mathcal{Q}$ is automatically labeled and, together with $\mathcal{L}$, forms the set of active images for the next cycle.
Before the next training cycle starts, we regard $\mathcal{R}=\mathcal{L}\cup\mathcal{Q}$ as the set of annotations which are potentially reviewed.
Inspired by~\cite{schubert2023identifying}, we introduce a post-processing label error detection method where the detection of \emph{misses} and \emph{flips} are two independent tasks.
When both types of label errors are simultaneously present, we use a parameter $\alpha\in[0,1]$ to distribute the review budget $\mathcal{C}_\mathcal{R}$ in reviewing \emph{misses} ($\alpha\mathcal{C}_\mathcal{R}$) and \emph{flips} ($(1-\alpha)\mathcal{C}_\mathcal{R}$).
In the following, we introduce two different review functions: a random review and a highest loss based review~\cite{schubert2023identifying}. 

\begin{figure*}
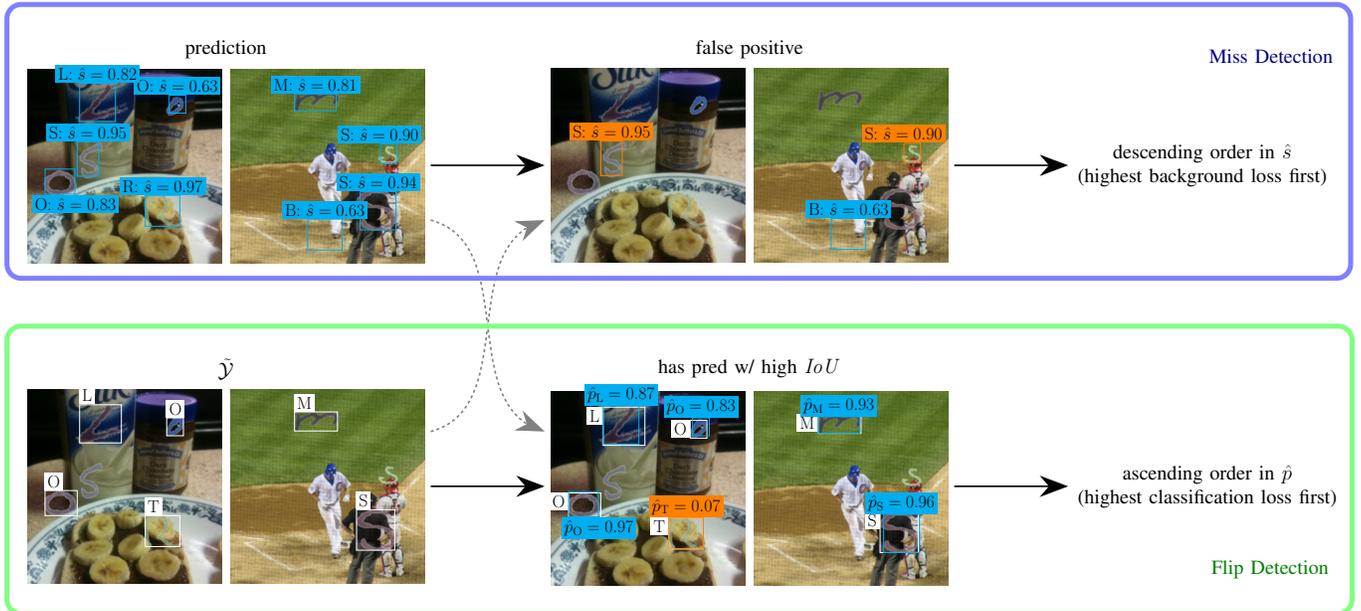

    \centering
    \resizebox{\linewidth}{!}{
        \includestandalone{figs/label_error_detection}
    }
    \caption{Schematic illustration of the label error detection mechanism using highest loss for missed labels (top, blue) and label flips (bottom, green).
    The detection of misses considers false positive predictions of the object detector while the detection of flips considers ground truth boxes, each of them matching at least one of the predictions in localization.
    }
    \label{fig:label-error-detection}
\end{figure*}
An illustration of our label error detection method can be found in \cref{fig:label-error-detection}.
We consider the set of all predictions on images from $\mathcal{R}$ and the corresponding noisy labels $\tilde{\mathcal{Y}}$.
To detect \emph{misses}, we select those predictions that are identified as false positive predictions according to the noisy ground truth $\tilde{\mathcal{Y}}$. 
To get an order for the review, we sort the false positives 
in descending order by the objectness score $\hat{s}$ for the highest loss based review and in random order for the random review.
Large values of $\hat{s}$ on false positives, where objectness is supposed to be small, amounts to a large objectness loss.

For the \emph{flips}, every label from $\tilde{\mathcal{Y}}$ is assigned to the most overlapping prediction if the $\iou$ of the two boxes is greater than of equal to $\iou_{\!\epsilon}$.
Then, the cross-entropy loss of the possibly incorrect label and the predicted probability distribution is used as a review score for every given label.
In case there is no sufficiently overlapping prediction, the respective label is not considered for review.
Note, that for given label class $\tilde{c}$, the cross-entropy loss of the assigned prediction $\hat{b}$ is 
\begin{equation}
    \mathrm{CE}(\hat{b} | \tilde{c}) = - \sum_{c = 1}^C \delta_{c \tilde{c}} \log\left(\hat{p}_c\right) = - \log\left(\hat{p}_{\tilde{c}}\right),
\end{equation}
where $\delta_{ij}$ is the Kronecker symbol, i.e., $\delta_{ij} = 1$ if $i = j$ and $\delta_{ij} = 0$ otherwise.
That is, the label with assigned prediction that has the lowest corresponding class probability $\hat{p}_{\tilde{c}}$ generates the highest loss.
In case of the random review method, we randomly select assigned labels for review by uniformly sampling over all  labels. 

\section{Numerical Experiments}
\noindent
In this section, we present our active learning setup with automatically labeling and reviewing data as well as new active learning hyperparameters.
Afterwards, we show results for both query functions with and without review for two different datasets and object detectors in terms of $\map$.
We also measure the performance of the review proposal mechanism in terms of precision.

\subsection{Experimental Setup}
\noindent
For our active learning setup, automated labeling and reviewing is desirable.
Therefore, we simulate label errors for all training images of the underlying datasets.
We do not include evaluations of the active learning experiments on the widely used MS-COCO or Pascal VOC datasets.
For an automated review procedure, the frequently occurring label errors in both datasets~\cite{schubert2023identifying} would lead to strongly biased results.
Evaluations on either dataset would require manual annotation review after each active learning cycle for several repetitions of the same experiment.
This manual review after each cycle is necessary in practice, however, out of scope for an experimental evaluation of the proposed method. 

\paragraph{Datasets and Models}
We make use of the EMNIST-Det dataset~\cite{riedlinger2022activelearning} with 20,000 training images and 2,000 test images as well as BDD100k~\cite{yu2020bdd100k} (BDD), where we filter the training and validation split, such that we only use daytime images with clear weather conditions, resulting in 12,454 training images.
Furthermore, the validation set is split into equally-sized test and validation sets, each consisting of 882 images.
Since EMNIST-Det is a simpler task to learn compared to BDD, we apply a RetinaNet~\cite{lin2017focal} and a Faster R-CNN~\cite{ren2015faster} with a ResNet-18~\cite{he_deep_2016} backbone for EMNIST-Det, as well as a Faster R-CNN with a ResNet-101 backbone for BDD.
Note, that this setup was introduced and used in related work~\cite{riedlinger2022activelearning,schubert2023identifying}.

Based on clean training data, the test performance for EMNIST-Det in terms of $\map$ is $91.2\%$ for Faster R-CNN and $90.9\%$ for RetinaNet.
For Faster R-CNN, the test performance decreases to $90.2\%$ with simulated \emph{misses} in the training data and to $89.2\%$ with simulated \emph{flips}.
Simulating \emph{misses} and \emph{flips} simultaneously yields a test performance of $89.4\%$ for Faster R-CNN and $89.3\%$ for RetinaNet.
For BDD and Faster R-CNN, a test performance of $50.0\%$ is obtained for unmodified training data and $48.9\%$ for training data including \emph{misses} and \emph{drops}.

\paragraph{Simulation of Label Errors}
\begin{figure}
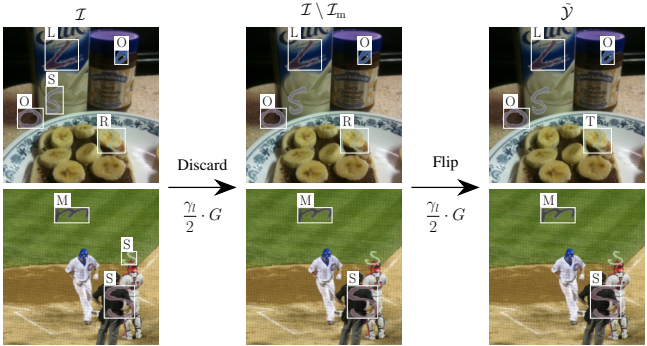

    \centering
    \resizebox{\linewidth}{!}{
        \includestandalone{figs/label_error_generation}
    }
    \caption{Schematic illustration of noise injection into a clean dataset.
        Misses are generated by randomly discarding $\gamma_l/2 \cdot G$ of the present annotations.
        Afterwards, $\gamma_l/2 \cdot G$ of the remaining annotations receive random class flips to one of the $C-1$ incorrect classes.
        This results in the noisy ground truth $\tilde{\mathcal{Y}}$ used in our experiments.
    }
    \label{fig:label-error-generation}
\end{figure}

\begin{table*}
    \centering
    \caption{Overview of important training, review and active learning hyperparameters for all datasets and networks.}
    \resizebox{\linewidth}{!}{
    \begin{tabular}{l l|c c c c|c c c c|c c}
         &  & \multicolumn{4}{c|}{Active Learning} & \multicolumn{4}{c|}{Review} & \multicolumn{2}{c}{Training} \\
         &  & $\vert \mathcal{U}_{init}\vert$ & $\mathcal{C}$ & $s_\epsilon$ & $T$ & $\gamma_l$ & $\gamma_r$ & $\iou_{\!\epsilon}$ & $\alpha$ & batch size & training iters \\
         \midrule
         Faster R-CNN & EMNIST-Det & $150$ & $200$ & $0.7$ & $20$ & $0.2$ & $0.05$ & $0.3$ & $0.5$ & $4$ & $25,\!000$ \\
         RetinaNet & EMNIST-Det & $150$ & $200$ & $0.5$ & $20$ & $0.2$ & $0.05$ & $0.3$ & $0.5$ & $4$ & $38,\!000$ \\
         \midrule
         Faster R-CNN & BDD100k & $625$ & $10,\!000$ & $0.7$ & $7$ & $0.2$ & $0.05$ & $0.3$ & $0.5$ & $4$ & $170,\!000$ \\
        \bottomrule
    \end{tabular}
    }
    \label{tab: al hyperparameters}
\end{table*}

For the simulation of \emph{misses} and \emph{flips}, we follow~\cite{schubert2023identifying}.
An illustration of the label error injection scheme can be found in \cref{fig:label-error-generation}.
Any dataset is equipped with a set of $G$ labels, i.e., 
\begin{equation}
    \mathcal{Y}=\{ b^i=(x^i, y^i, w^i, h^i, c^i ):i=1,\ldots,G \}.
\end{equation}
Let $\mathcal{I} = \{ 1,\ldots,G \}$ be the set of indices of all boxes $b^i\in\mathcal{Y},\, i=1,\ldots,G$.
Furthermore, we choose a parameter $\gamma_l \in [0,1]$ representing the relative frequency of label errors during data acquisition.

For generating label misses, we randomly choose a subset $\mathcal{I}_\text{m} \subset \mathcal{I}$ of size $\frac{\gamma_l}{2}\cdot G$, representing missed ground truth boxes which are discarded from $\mathcal{Y}$.

The number of remaining annotations that receive a class flip is again $\frac{\gamma_l}{2}\cdot G$, where the class flip is determined uniformly over the $C - 1$ incorrect class assignments.
The potentially flipped class for each label $b^i$ is denoted by $\tilde{c}^i$.
Finally, we denote the training set including label errors by 
\begin{equation}
    \tilde{\mathcal{Y}} = \{ (x^i, y^i, w^i, h^i, \tilde{c}^i) : i \in \mathcal{I}\setminus\mathcal{I}_\text{m}\}.
\end{equation}

Note, that a single label is perturbed by only one type of label error at most.
In addition, label errors are not simulated on the test dataset to ensure an unbiased evaluation of test performance. 

\paragraph{Automated Review of Label Errors}
Since the  oracle is noisy with error frequency $\gamma_r$, the review is also error-prone, i.e., \emph{misses} are detected with probability $1-\gamma_r$ and still overlooked with probability $\gamma_r$.
The \emph{flips}, whether the label error proposal was a false alarm or not, are corrected with probability $1-\gamma_r$ and randomly misclassified with probability $\gamma_r$.

\begin{figure}
    \centering
    \includegraphics[trim={0.5cm 0.45cm 0.5cm 0.5cm},clip,width=\linewidth]{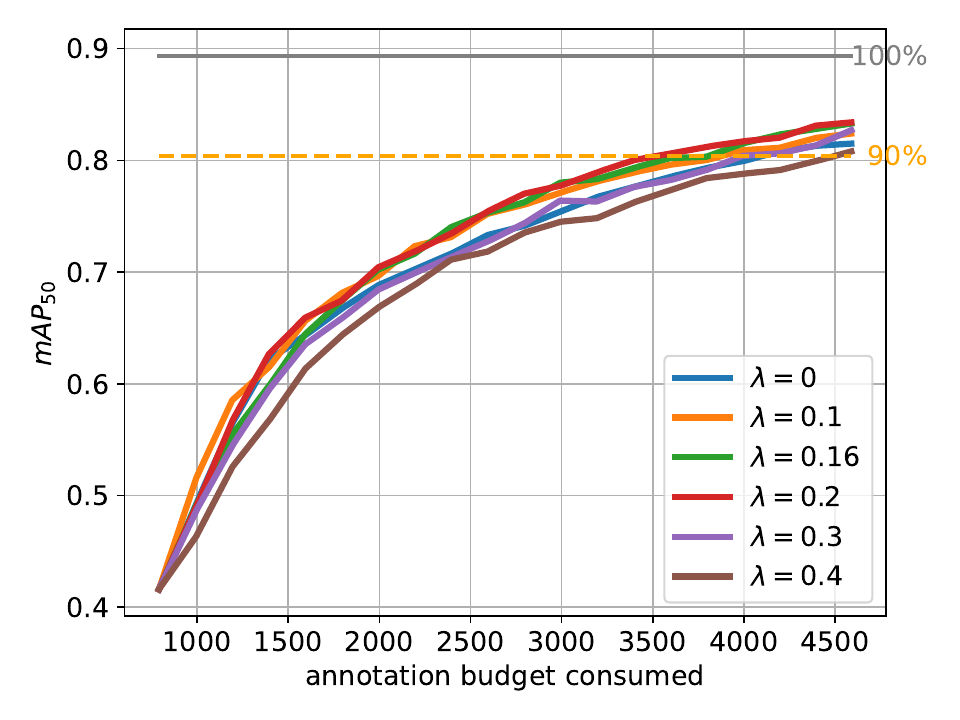}   
    \caption{EMNIST-Det ablation study of the ratio between labeled and reviewed bounding boxes for Faster R-CNN where both label error types are present. We compare the random query without review with random query and highest loss review (RHL) with the chosen ration $\lambda$ in brackets.}
    \label{fig: ablation lambda}
\end{figure}

\begin{figure*}
    \centering
    \begin{subfigure}[c]{0.32\textwidth}
    \includegraphics[trim={0.6cm 0.55cm 0.6cm 0.5cm},clip,width=\textwidth]{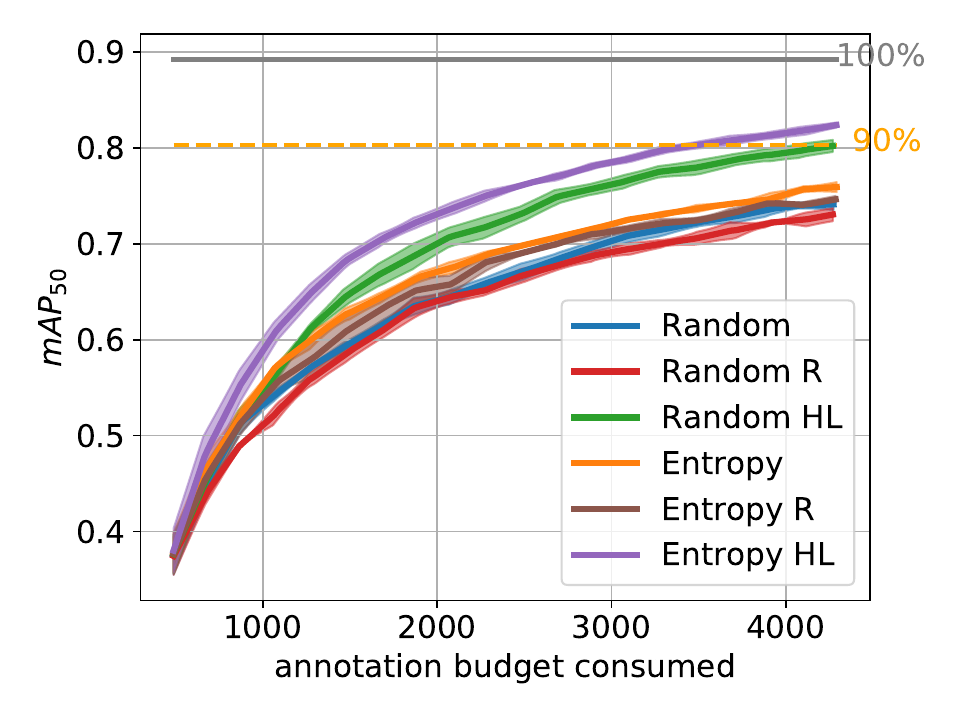}
    \subcaption{Class Flips}
    \end{subfigure}
    \begin{subfigure}[c]{0.32\textwidth}
    \includegraphics[trim={0.6cm 0.55cm 0.6cm 0.5cm},clip,width=\textwidth]{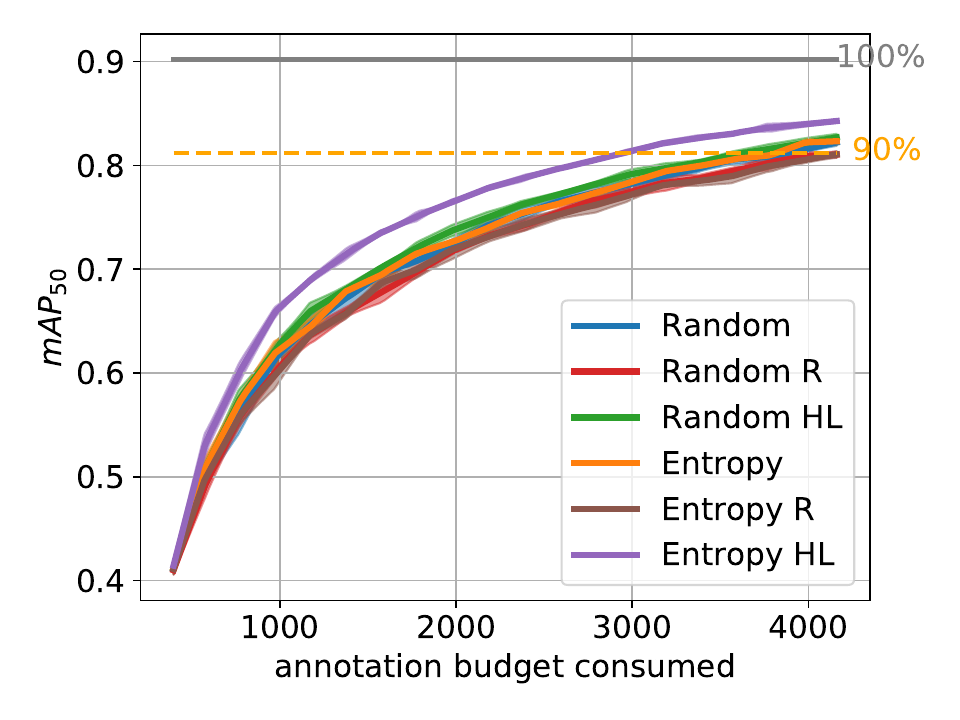}
    \subcaption{Label Misses}
    \end{subfigure}
    \begin{subfigure}[c]{0.32\textwidth}
    \includegraphics[trim={0.6cm 0.55cm 0.6cm 0.5cm},clip,width=\textwidth]{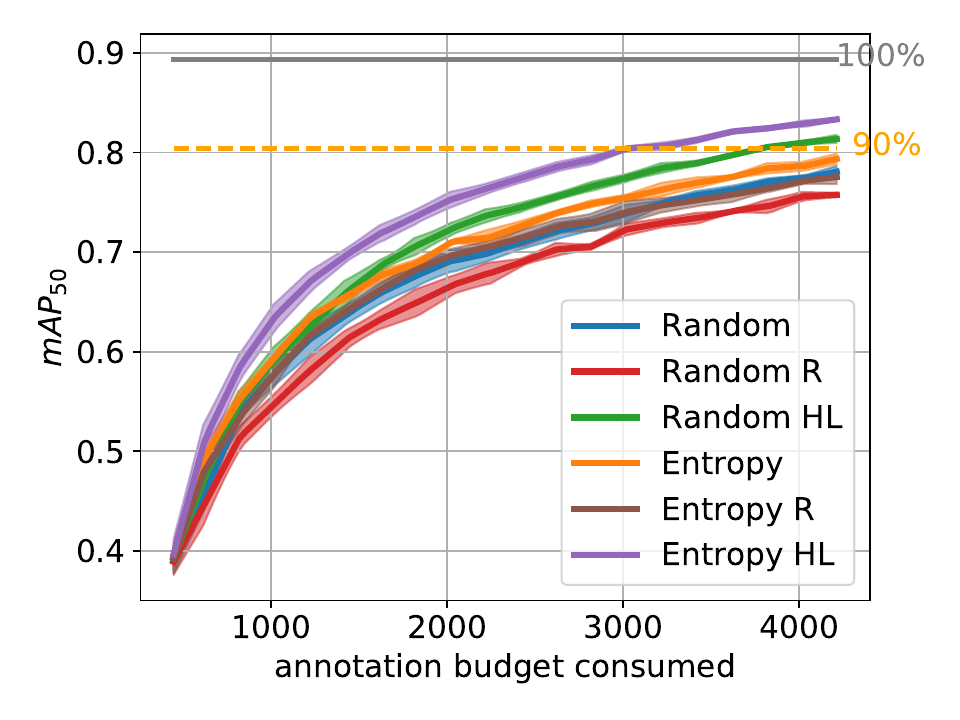}
    \subcaption{Flips + Misses}
    \end{subfigure}
    \caption{EMNIST-Det active learning curves, where only flips are present (left), only misses (center) as well as where flips and misses are simultaneously present (right).}
    \label{fig: EMNIST active learning results}
\end{figure*}

\paragraph{Implementation Details}
We implemented our active learning methods in the open source MMDetection toolbox~\cite{mmdetection}.
For the label error simulation, we choose the relative frequency of label errors $\gamma_l=0.2$, the relative frequency of label errors during review $\gamma_r=0.05$, the value for score-thresholding $s_\epsilon=0.7$ for Faster R-CNN and $s_\epsilon=0.5$ for RetinaNet as well as the $\iou$-value $\iou_{\!\epsilon} = 0.3$ that assigns predictions with labels. 
We choose $\gamma_r<\gamma_l$, assuming that the  oracle is more engaged in viewing and evaluating single boxes during the review compared to labeling from scratch, with all boxes having to be located and classified on a new image.
As hyperparameters for the active learning cycle, the initially labeled set consists of 150 randomly picked images EMNIST-Det and of 625 randomly picked images for BDD.
The budget for a single active learning step $\mathcal{C}$ is 200 for EMNIST-Det and $\mathcal{C}=10,\!000$ for BDD.
Labeling a single box has cost $1$, as does reviewing a label error proposal, whether \emph{miss} or \emph{flip} and also whether a label error was identified or not.
If \emph{misses} and \emph{flips} are simultaneously present in the experiment, we set the ratio between reviewing \emph{misses} and \emph{flips} $\alpha=0.5$.
Finally, the number of active learning steps for EMNIST-Det is $T=20$ and for BDD $T=7$.
For an overview of training, review and active learning hyperparameters, see \cref{tab: al hyperparameters}.

\subsection{Results}
\noindent
In the following, we show active learning results for EMNIST-Det and BDD.
Therefore, we compare six different methods, the random and entropy query, both without review, as well with random review or review by highest loss.
Furthermore, we present performance results for both review methods in terms of precision over the whole active learning course.

\begin{figure*}
    \centering
    \begin{subfigure}[c]{0.32\textwidth}
    \includegraphics[trim={0.6cm 0.6cm 0.5cm 0.05cm},clip,width=\textwidth]{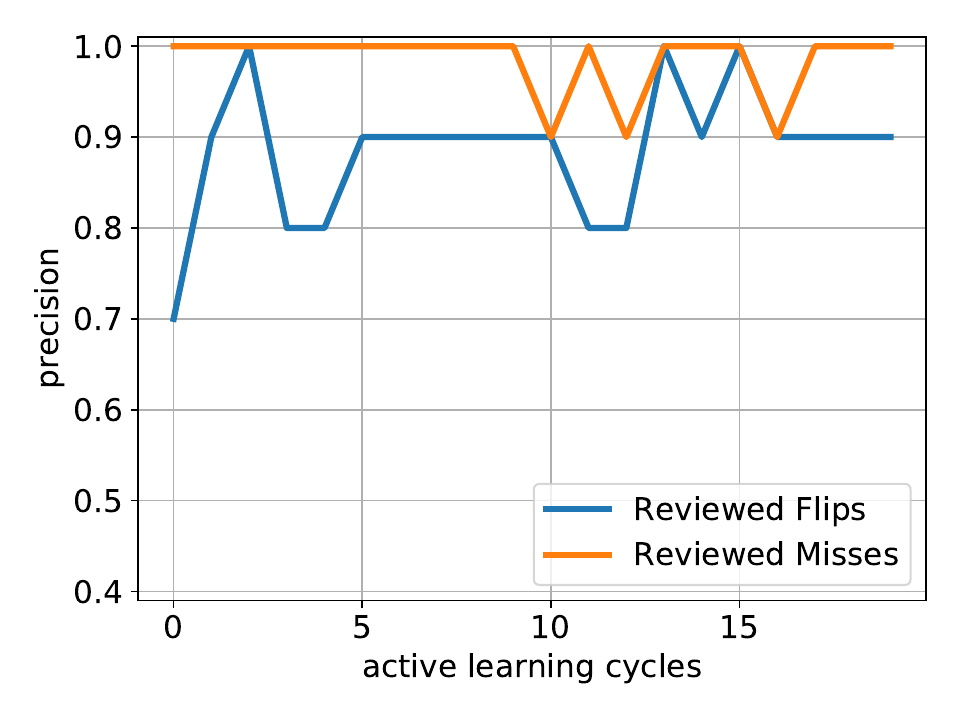}
    \subcaption{Faster R-CNN + EMNIST-Det}
    \end{subfigure}
    \hfill
    \begin{subfigure}[c]{0.32\textwidth}
    \includegraphics[trim={0.6cm 0.6cm 0.5cm 0.05cm},clip,width=\textwidth]{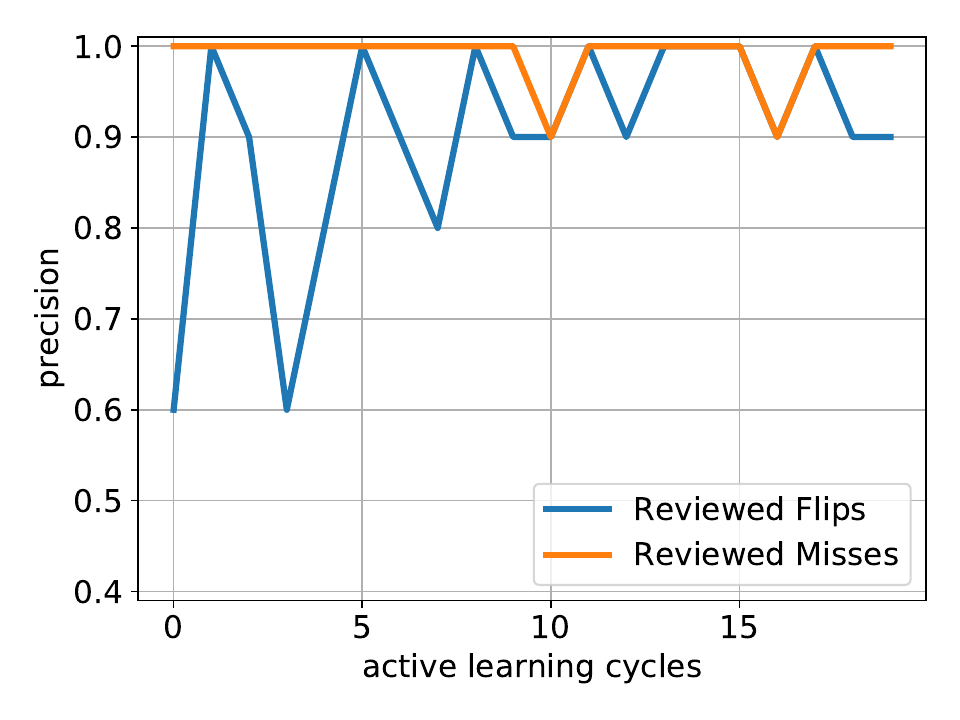}
    \subcaption{RetinaNet + EMNIST-Det}
    \end{subfigure}
    \hfill
    \begin{subfigure}[c]{0.32\textwidth}
    \includegraphics[trim={0.6cm 0.6cm 0.5cm 0.05cm},clip,width=\textwidth]{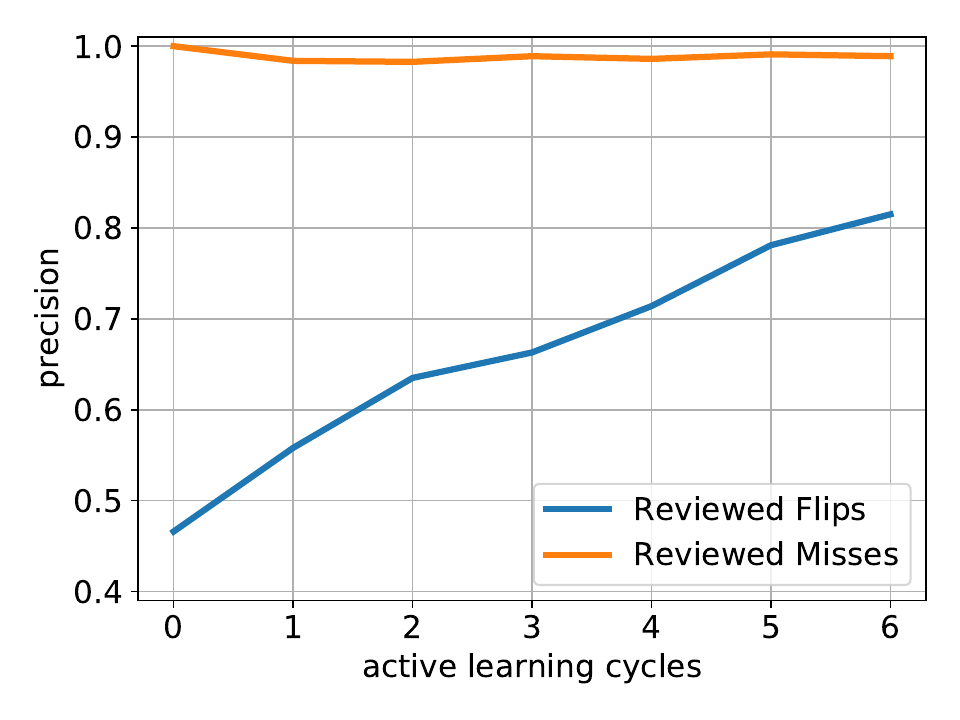}
    \subcaption{Faster R-CNN + BDD}
    \end{subfigure}
    \\
    \begin{subfigure}[c]{0.32\textwidth}
    \includegraphics[trim={0.6cm 0.6cm 0.5cm 0.05cm},clip,width=\textwidth]{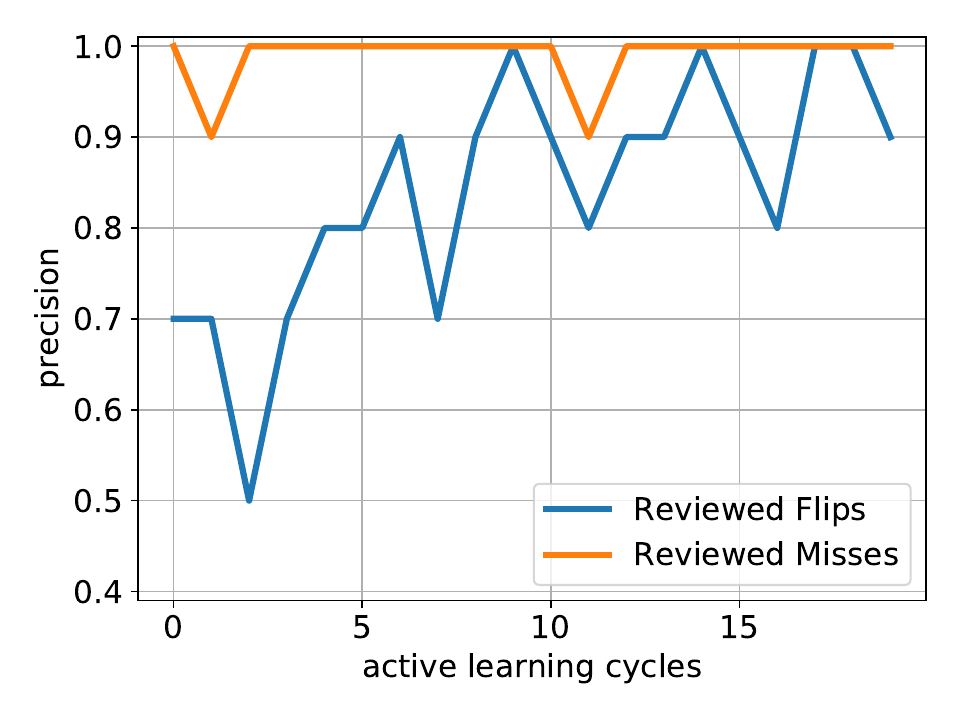}
    \subcaption{Faster R-CNN + EMNIST-Det}
    \end{subfigure}
    \hfill
    \begin{subfigure}[c]{0.32\textwidth}
    \includegraphics[trim={0.6cm 0.6cm 0.5cm 0.05cm},clip,width=\textwidth]{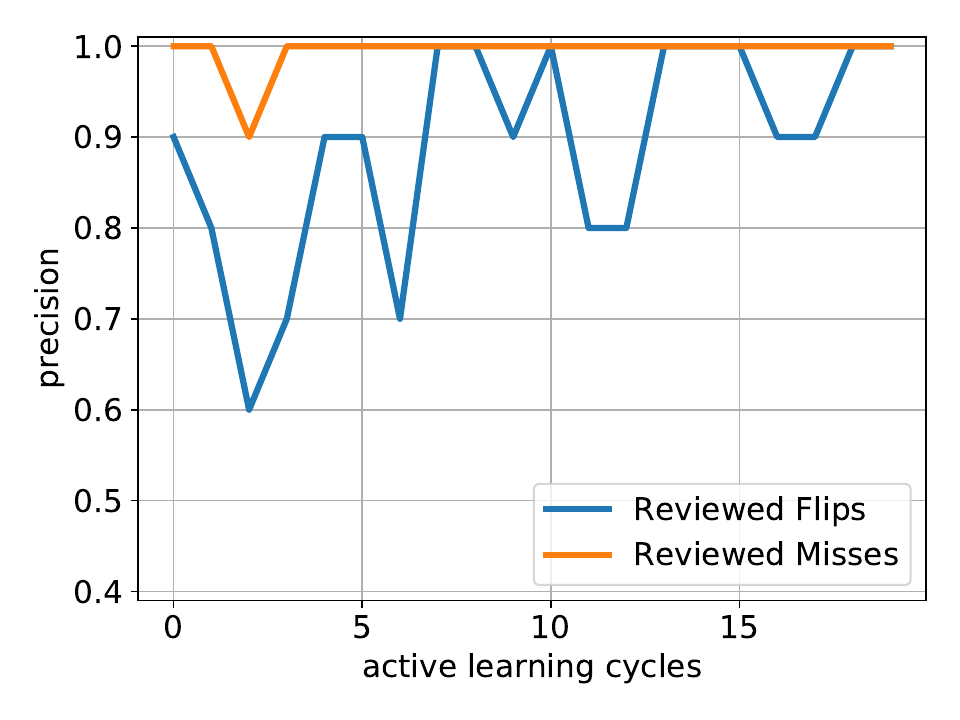}
    \subcaption{RetinaNet + EMNIST-Det}
    \end{subfigure}
    \hfill
    \begin{subfigure}[c]{0.32\textwidth}
    \includegraphics[trim={0.6cm 0.6cm 0.5cm 0.05cm},clip,width=\textwidth]{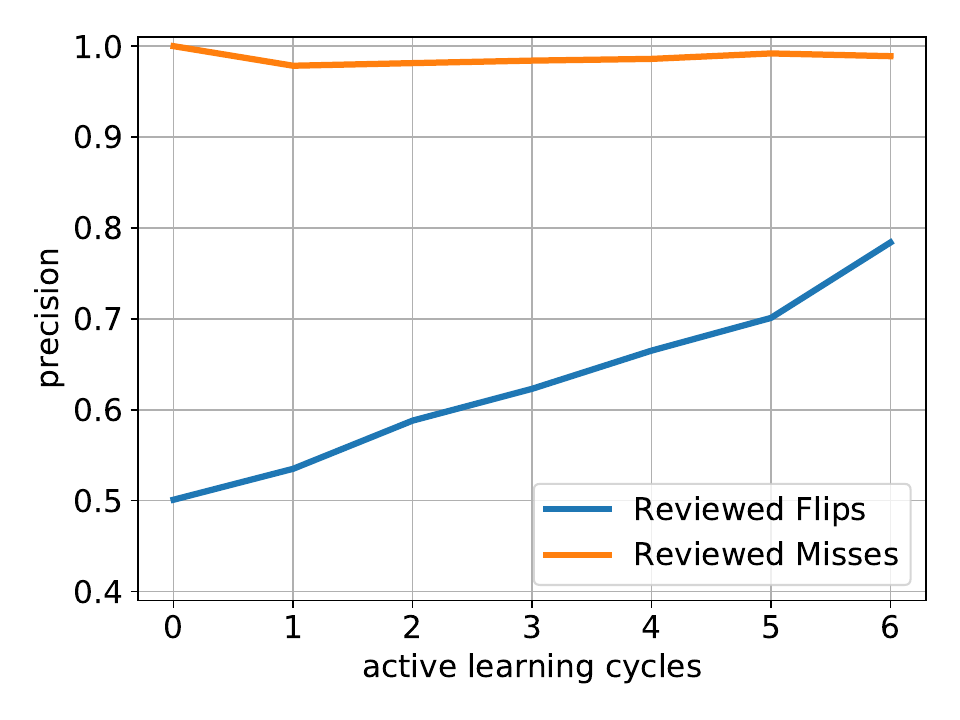}
    \subcaption{Faster R-CNN + BDD}
    \end{subfigure}
    \caption{Review quality results for random highest loss (top) and entropy highest loss (bottom). Misses and flips are simultaneously present in all experiments.}
    \label{fig: review quality}
\end{figure*}

\begin{figure*}
    \centering
    \begin{subfigure}[c]{0.48\textwidth}
    \includegraphics[trim={0.5cm 0.45cm 0.5cm 0.5cm},clip,width=\textwidth]{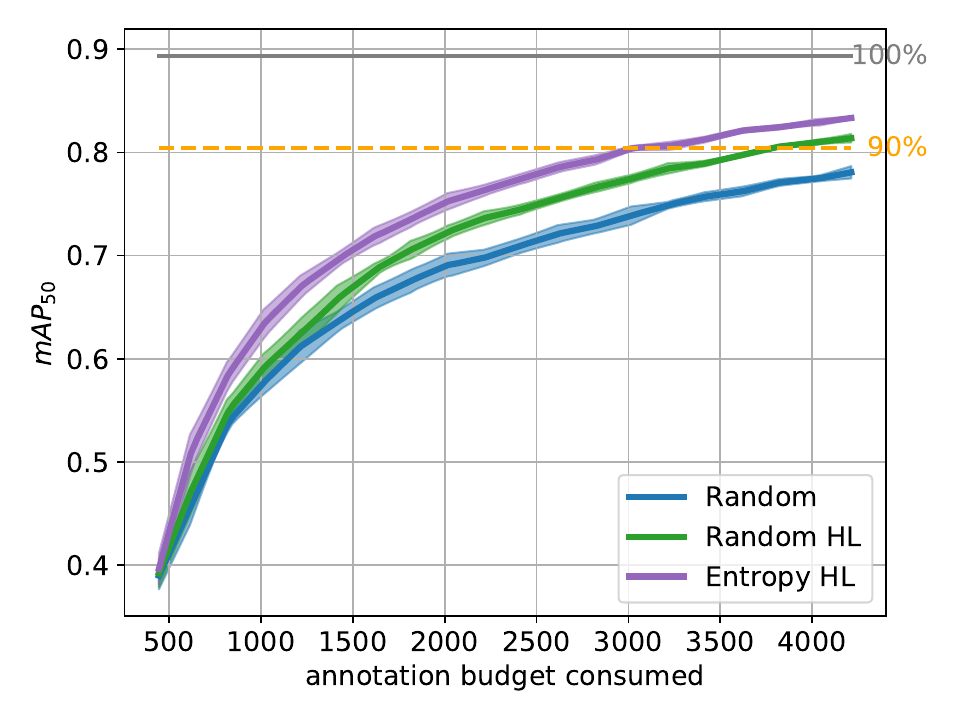}
    \subcaption{Faster R-CNN}
    \end{subfigure}
    \begin{subfigure}[c]{0.48\textwidth}
    \includegraphics[trim={0.5cm 0.45cm 0.5cm 0.5cm},clip,width=\textwidth]{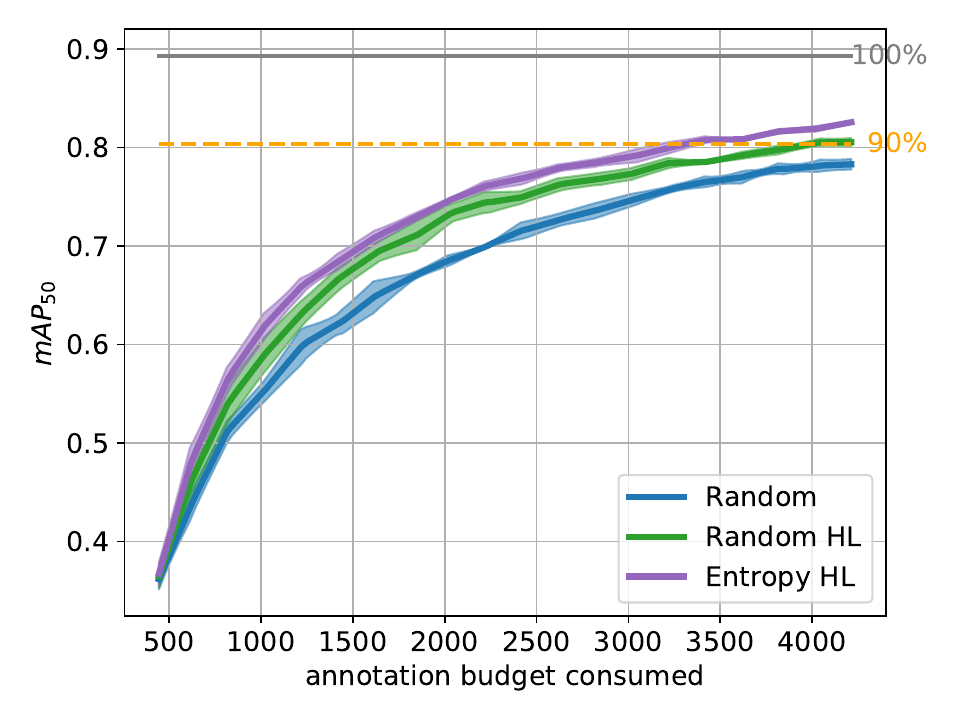}
    \subcaption{RetinaNet}
    \end{subfigure}
    \caption{Comparison of EMNIST-Det active learning curves for Faster R-CNN (left) and RetinaNet (right) where both label error types are present.}
    \label{fig: EMNIST active learning results comparison}
\end{figure*}

\noindent
\paragraph{Ablation for the Ratio of Queried and Reviewed Bounding Boxes}
For the methods with review, the fraction $\lambda$ of the amount of new data queries and the amount of bounding box reviews plays a significant role.
Therefore, we repeat the same experiment for EMNIST-Det and Faster R-CNN with different values for $\lambda$, see \cref{fig: ablation lambda}.
These results are based on training data with simulated \emph{flips} and \emph{misses}.
The gray and yellow lines indicate the $100\%$ and $90\%$ reference performance mark of the model trained with the entire (noisy) dataset.
The random query with highest loss review (RHL) is visually almost identical for $\lambda=0.1$, $\lambda=0.16$ and $\lambda=0.2$. 
All these three methods outperform the random query without review.
The random query without review performs similar to the random query with highest loss review with $\lambda=0.3$ and outperforms the random query with $\lambda=0.4$, i.e., at about $\lambda=0.3$ is the break-even-point, at which it is no longer worthwhile to review more bounding boxes instead of labeling new ones.
We hypothesize that this tipping point is strongly dependent on our chosen setup with a relative frequency of label errors of $\gamma_l=0.2$.
Since the red curve seems to be most favorable, we set the fraction between queries and reviews to $\lambda=0.2$ in all of the following experiments.

\paragraph{Active Learning with Different Label Error Types}
We first investigate active learning curves for EMNIST-Det and Faster R-CNN.
We consider active learning curves where a) we simulated only \emph{misses}, b) only \emph{flips}, and c) both label error types occur equally often in the training dataset, each with noise rate $\gamma_l=0.1$ (recall \cref{fig:label-error-generation}).
We compare both query strategies, random and entropy, without review, with highest loss (HL) review and with random (R) review, respectively.
The obtained active learning curves are averaged over four random initializations and evaluated in terms of the total annotation budget consumed.
Note, that for the active learning methods without review the total amount of annotation budget is equal to the amount of (possibly incorrect) labeled bounding boxes.
For those methods incorporating a review step the total amount of annotation budget represents the sum of labeled and reviewed bounding boxes.
\Cref{fig: EMNIST active learning results} shows active learning curves in terms of test performance with point-wise standard deviations.
For all three active learning curves, we observe that the entropy method outperforms random at every point.
Furthermore, the queries without review perform superior to the respective query with random review.
Entropy HL and random HL, i.e., both methods with highest loss review clearly outperform the strategies without review and with random review.
We conclude that the success of reviewing queries strongly depends on the performance of the review methods and that random review is too expensive in terms of annotation budget.
From this we conclude that it is more worthwhile to acquire new (noisy) labels than to randomly review the active labels, at least for the given amount of noise we studied.

All in all, the distance between the active learning curves of the six methods is significantly larger for label \emph{flips} as compared to \emph{misses}.
Moreover, the maximum performance with simulated label errors is also inferior for the \emph{flips} compared to the \emph{misses}.
We hypothesize that the reason for this is the sub-sampling from the negatively associated anchors~\cite{ren2015faster} during training.
This mechanism leads to only partial learning from the \emph{misses}, whereas an incorrect foreground class induced from \emph{flips} has a negative impact on every gradient step.

The significant difference of the active learning curves of the respective queries with random review and the highest loss review can be attributed to the high precision of the highest loss review.
The random review has an expected precision of $\gamma_l$.
\Cref{fig: review quality} shows the precision for the highest loss review applied after random query in (a) and after entropy query in (d) across the span of all active learning cycles.
In both plots, (a) and (d), \emph{flips} and \emph{misses} are simultaneously present, i.e., both plots correspond to the respective method from \cref{fig: EMNIST active learning results} (c).
The blue lines visualize the precision for the review identifying a \emph{flip} and the orange lines analogously for the \emph{misses}.
Here, the precision for detecting \emph{flips} is always above $50\%$ and tends to improve as the active learning experiment progresses, whereas the precision for the detection of \emph{misses} is even consistently above $90\%$.
In general, \emph{flips} are more difficult to detect compared to \emph{misses} due to the different construction of the detection methods of either label error type, recall \cref{fig:label-error-detection}.

\begin{table}
\centering
 \caption{Mean average precision values in $\%$ with standard deviations in brackets for 2000 and 4000 queried and reviewed annotations for Faster R-CNN (top) and RetinaNet (bottom) for EMNIST-Det. 
 Note, that in every experiment both label error types are present; the upper half represents \cref{fig: EMNIST active learning results comparison} a) and the bottom half \cref{fig: EMNIST active learning results comparison} b).
 } 
\resizebox{\linewidth}{!}{
 \begin{tabular}{c c c c}
 Network & Method & $\map_{\!@2000}$ & $\map_{\!@4000}$ \\
 \toprule
 \multirow{3}{*}{Faster R-CNN} & Random & $68.97(\pm 1.09)$ & $77.41(\pm 0.28)$ \\
  & Random HL & $72.13(\pm 0.64)$ & $80.92(\pm 0.23)$ \\
  & Entropy HL & $75.13(\pm 0.79)$ & $82.87(\pm 0.29)$ \\
 \midrule
 \multirow{3}{*}{RetinaNet} & Random & $68.38(\pm 0.52)$ & $78.04(\pm 0.51)$ \\
  & Random HL & $72.94(\pm 1.06)$ & $80.40(\pm 0.34)$ \\
  & Entropy HL & $74.39(\pm 0.28)$ & $81.86(\pm 0.27)$ \\
 \bottomrule
 \end{tabular}
 }
 \label{tab: al results emnist}
\end{table}

\paragraph{Comparing Faster R-CNN with RetinaNet on EMNIST-Det}
For the following results, we compare only the random query without review with the random and entropy query, both with highest loss review.
\Cref{fig: EMNIST active learning results comparison} shows active learning curves for these methods for Faster R-CNN in (a) and for RetinaNet in (b).
Note, that in both cases both label error types are present.
Moreover, (a) is a trimmed version of \cref{fig: EMNIST active learning results} (c) to make it more convenient to compare the results from both detectors visually.
We observe that the curves for Faster R-CNN and RetinaNet look very similar over the entire active learning course.
All curves start at just below $40\%$ $\map$ and the respective methods end at similar test performances.
The ranking of the methods is always the same: entropy HL outperforms random HL and random without review.
Also, random HL outperforms random without review.
For RetinaNet, random HL seems to be closer to entropy HL as compared to Faster R-CNN.

These observations are also supported by \cref{tab: al results emnist}, wherein we stated the $\map$ values with standard deviations in parentheses.
We compare performance for the total annotation budget consumed equal to $2000$ and $4000$ from the active learning curves shown in \cref{fig: EMNIST active learning results comparison}.
In particular, for entropy with highest loss review, the $\map_{\!@2000}$ for Faster R-CNN is $0.74$ percent points (pp) higher and even $1.01$ $\map_{\!@4000}$ pp higher.
For Faster R-CNN, the difference between entropy HL and random HL is $3$ pp for $\map_{\!@2000}$ and $1.95$ for $\map_{\!@4000}$.
For RetinaNet, the difference is only $1.45$ pp for $\map_{\!@2000}$ and $1.46$ for $\map_{\!@4000}$.

Comparing the quality of the highest loss review, the results for RetinaNet are highly correlated to the results of Faster R-CNN, see \cref{fig: review quality}.
For RetinaNet, the precision for the highest loss review in combination with random query is visualized in (b) and with the entropy query in (e).
Here, the precision for detecting the \emph{misses} is at or above $90\%$.
The precision for the detection of \emph{flips} is always greater or equal to $60\%$ and from active learning cycle $7$ onward even always above $80\%$.
We observe that the precision of the highest loss review increases while the experiments progresses, i.e., object detectors trained with more data generate better label error proposals.
We hypothesize that with more data, overfitting can be more effectively prevented and that the object detectors will generalize better, thus label errors in the active labels will not be as significant when sufficient data is available.

\begin{figure}
    \centering
    \includegraphics[trim={0.5cm 0.45cm 0.5cm 0.5cm},clip,width=\linewidth]{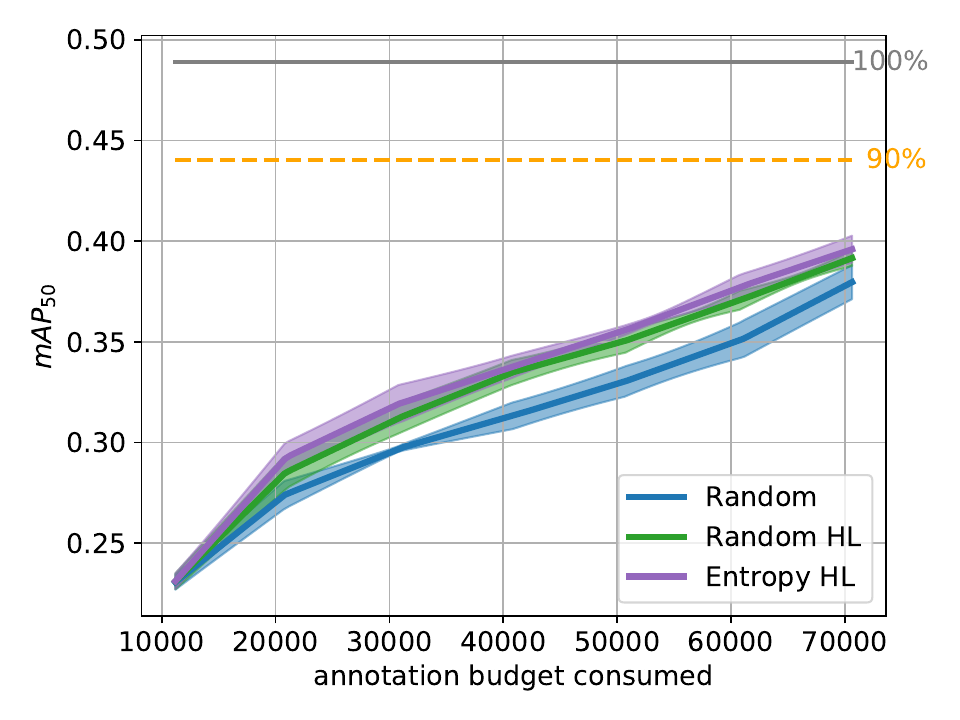}    
    \caption{BDD active learning curves for Faster R-CNN where both label error types are present.}
    \label{fig: BDD active learning results}
\end{figure}

\paragraph{Results for BDD with Faster R-CNN}
\Cref{fig: BDD active learning results} shows active learning curves for the random query without review, as well as for the random and entropy query both with highest loss review.
Comparable to the results for EMNIST-Det, the ranking of the methods is identical over the entire active learning course.
The random query without review is inferior to both queries with review and entropy HL is superior to random HL.
Note, that the distance between the two queries with review is marginal.
In contrast, there is a significant difference between either one and the random query without review.

The review quality of the highest loss review for the random query is shown in \cref{fig: review quality} (c) and for the entropy query in (f).
Again, the \emph{misses} are detected at all times with a precision of nearly $100\%$.
Starting at just under $50\%$, the precision for identifying \emph{flips} increases steadily over the active learning course ending at close to $80\%$.
We conclude that involving a label review in the active learning cycle is also highly beneficial in the more complex BDD real world dataset.
Analogous to the results for EMNIST-Det, the highest loss review becomes more precise as the experiments progress and the number of active labels increases.

\section{Conclusion}
\noindent
In this work, we considered label errors in active learning cycles for object detection for the first time, where we assumed a noisy oracle during the annotation process.
We realized this assumption by simulating two types of label errors for the training data of datasets which are reasonably free of intrinsic label errors.
These types of label errors are missing bounding box labels as well as bounding box labels with an incorrect class assignment.
We introduce a review module to the active learning cycle, that takes as input the currently labeled images and the corresponding predictions of the most recently trained object detector.
Furthermore, we detect both types of label errors by a random review method and a method based on the highest loss of the model's predictions and the corresponding noisy labels.
We observe that the incorporation of random review leads to an even worse test performance compared to the corresponding query without review.
Nevertheless, we show that the combination of query strategies, like random selection or instance-wise entropy, with an accurate review yields a significant performance increase.
For both query strategies, the improvement obtained by including the highest loss review persists during the whole active learning course for different dataset-network-combinations.
We make our code for reproducing results and further development publicly available at \texttt{GitHub}.

\paragraph*{Acknowledgements}
We gratefully acknowledge financial support by the German Federal Ministry of Education and Research in the scope of ``Projekt UnrEAL'', grant no.\ 01IS22069.
The authors gratefully acknowledge the \href{https://www.gauss-centre.eu/}{Gauss Centre for Supercomputing e.V.} for funding this project by providing computing time through the Johnvon Neumann Institute for Computing on the GCS Supercomputer JUWELS at Julich Supercomputing Centre.

\bibliographystyle{plain}
\bibliography{biblio}

\end{document}